\documentclass[10pt,twocolumn,letterpaper]{article}

\usepackage[pagenumbers]{cvpr} 

\definecolor{cvprblue}{rgb}{0.21,0.49,0.74}
\usepackage[pagebackref,breaklinks,colorlinks,allcolors=cvprblue]{hyperref}

\usepackage{booktabs}
\usepackage{tabularx}

\definecolor{attackrole}{HTML}{B5523B}    

\usepackage[T1]{fontenc}        
\usepackage{upquote}            
\usepackage[most]{tcolorbox}

\definecolor{promptframe}{HTML}{3A5A8C}   
\definecolor{promptback}{HTML}{F5F7FB}    
\definecolor{sysrole}{HTML}{6B7A8F}       
\definecolor{userrole}{HTML}{2E7D5B}      
\definecolor{slotcol}{HTML}{C0392B}       

\lstdefinestyle{promptmsg}{
  basicstyle=\ttfamily\scriptsize, columns=fullflexible, keepspaces=true,
  breaklines=true, breakatwhitespace=true, frame=none,
  aboveskip=0pt, belowskip=0pt, xleftmargin=0pt, xrightmargin=0pt,
  moredelim=**[is][\color{slotcol}\bfseries]{@}{@}
}

\newtcblisting{chatmsg}[2]{
  enhanced, listing only, listing options={style=promptmsg},
  colback=white, colframe=#2, boxrule=0.5pt, arc=2pt,
  left=4pt, right=4pt, top=7pt, bottom=3pt, before skip=5pt, after skip=3pt,
  attach boxed title to top left={xshift=5pt, yshift=-2.2mm},
  boxed title style={colback=#2, colframe=#2, boxrule=0pt, arc=1pt,
                     left=2pt, right=2pt, top=0.5pt, bottom=0.5pt},
  coltitle=white, fonttitle=\sffamily\bfseries\tiny,
  title={\MakeUppercase{#1}}
}

\newtcolorbox{promptpanel}[2][]{
  enhanced, colback=promptback, colframe=promptframe,
  boxrule=0.6pt, arc=2pt, left=4pt, right=4pt, top=2pt, bottom=4pt,
  fonttitle=\sffamily\bfseries\footnotesize, title={#2},
  segmentation style={dashed, promptframe!40}, lower separated=true,
  fontlower=\footnotesize, #1
}
\newcommand{\slotname}[1]{\textcolor{slotcol}{\ttfamily\bfseries\{#1\}}}

\newcommand{\NA}{--}

\def\paperID{*****} 
\def\confName{CVPR}
\def\confYear{2027}

\title{Query-Guided Visual Minimization for Selective Disclosure to Cloud VLMs}

\newcommand\blfootnote[1]{%
  \begingroup
  \renewcommand\thefootnote{}\footnote{#1}%
  \addtocounter{footnote}{-1}%
  \endgroup
}

\author{
\begin{tabular}{c}
{\large
\textbf{Pedram Zaree}$^{1}$ \quad
\textbf{Vandana Malayil}$^{2}$ \quad
\textbf{Kleomenis Katevas}$^{2}$ \quad
\textbf{Hamed Haddadi}$^{2}$ \quad
\textbf{Nael Abu-Ghazaleh}$^{1}$ 
}
\\[0.5em]
$^{1}$CSE Department, University of California, Riverside, USA\\
$^{2}$Brave, London, United Kingdom
\end{tabular}
}

\begin{document}
\maketitle

\blfootnote{1. Work done during an internship at Brave.}
\blfootnote{2. Code will be released upon acceptance.}

\begin{abstract}
Cloud-hosted vision-language models (VLMs) require sending images beyond the user's local trust boundary, potentially exposing sensitive information the query does not need. We study query-guided visual minimization: a trusted local transformation preserves query-relevant visual content and reduces unnecessary disclosure before transmission. Unlike query-agnostic anonymization, it preserves the same sensitive content when the query requires it and suppresses it otherwise. We introduce \textbf{QGVM}, a framework that combines query grounding with an independent sensitive-item inventory, determines disclosure necessity at the type, instance, and text-content levels, and renders a minimized image for a black-box downstream VLM. We evaluate QGVM on complementary benchmarks measuring task utility, directly recoverable sensitive information, and inferential geolocation leakage.
With OpenAI GPT-6.1 Sol, QGVM achieves 85.3\% DoxBench answer correctness versus 77.5\% for PGP, while a separately prompted Sol geolocation attacker succeeds within 1\,km on 5.8\% of QGVM images versus 25.7\% of unprotected images. On Scene-Privacy, QGVM reduces face re-identification from 27.7\% (PGP) to 4.5\%.
On held-out BIV-Priv-Seg, unnecessary private-text recovery falls from 33.9\% (PGP) to 15.3\%.
To verify that disclosure is query-dependent, we evaluate paired queries that either require or do not require the same sensitive information. QGVM achieves 69.6\% correct disclosure switching, compared with 29.1\% for grounding-only cropping.
These findings highlight the value of distinguishing visual relevance from disclosure necessity, while revealing the privacy--utility trade-offs of selective visual disclosure.
\end{abstract}

\begin{figure}[!t]
\centering
\includegraphics[width=\columnwidth]{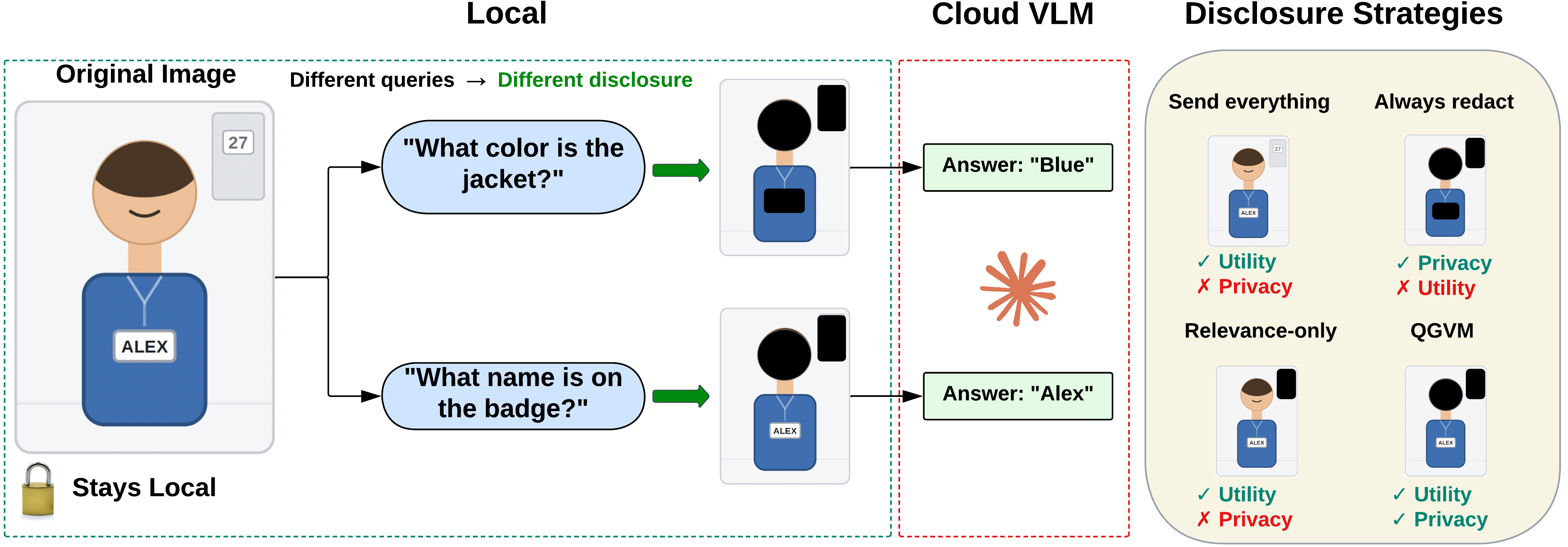}
\caption{\textbf{Query-guided visual minimization.}
The same image requires different disclosure depending on the query: identifying a jacket's color does not require revealing the wearer's face or badge, whereas reading the badge requires preserving its content. QGVM selectively protects unnecessary sensitive information, including within query-relevant regions, before cloud inference. The right panel contrasts QGVM with full-image transmission, query-agnostic redaction, and relevance-only selection.}
\label{fig:overview}
\end{figure}

\section{Introduction}
\label{sec:intro}

Vision-language models (VLMs) enable powerful visual question answering and scene understanding applications~\citep{antol2015vqa,singh2019towards,mathew2021docvqa, liu2023visual, alayrac2022flamingo, li2023blip}. Popular vision-enabled services, such as ChatGPT~\citep{hurst2024gpt}, Gemini~\citep{comanici2025gemini}, and Claude~\citep{anthropic2026opus55}, rely on cloud-hosted models. Image queries to them typically send visual content beyond the user's local trust boundary, potentially exposing private information unrelated to the query. Everyday images may contain sensitive information, including identities, documents, screenshots, locations, and bystanders. Even seemingly innocuous contextual cues, such as background landmarks, can also reveal sensitive attributes or geographic locations~\citep{tomekcce2024private,luo2025doxing}.

We quantify this exposure using 256 image--query cases over 149 real-world images from DoxBench~\citep{luo2025doxing}.
For this analysis, we constructed task queries targeting non-geographic visual information, such as object appearance, rather than the location depicted in the image. With Claude Opus~5.5, unprotected images achieve 82.2\% answer correctness, while a separately prompted Opus geolocation attacker localizes 58.3\% of cases within a 25\,km radius of their ground-truth locations. Scene-Privacy further examines images containing sensitive information that may or may not be required by the query. For queries where the targeted information is unnecessary, unprotected transmission permits 98.9\% face re-identification and 80.2\% private-text recovery. These results illustrate the gap between the visual content needed for a task and the unnecessary private information exposed through full-image transmission.

This raises a central question: \emph{Can we reduce visual privacy exposure by transmitting only the content needed to answer the user's query?} Many visual queries require only a subset of the image: for example, recognizing a plant species may require only the plant itself and not the background showing a child playing. Removing visual content based on predefined privacy labels, such as a face, is not sufficient. Instead, what should be disclosed must depend on the query. A name badge, for example, may be necessary when asked to read it but unnecessary when asked about the wearer's clothing (Fig.~\ref{fig:overview}). This task-dependent disclosure principle is motivated by \emph{contextual integrity}~\citep{nissenbaum2004privacy}, which considers the appropriateness of information flow within a given context, and \emph{data minimization}~\citep{tankard2016gdpr}, which limits disclosure to information necessary for a specified purpose. We therefore seek to preserve query-required visual content while minimizing what is unnecessarily revealed. We assume that sensitive content required by the query is eligible for disclosure; determining whether such disclosure is authorized or subject to additional privacy preferences is outside our scope.

We study query-guided visual minimization in a local-to-cloud setting: a trusted local component transforms the image before transmission, without querying or accessing the parameters of the downstream VLM. Our motivation behind choosing this setting is as follows: Mobile visual querying is already widespread; for example, Google reports over 20 billion visual searches through Lens~\citep{googlelens} each month, including camera-based queries on smartphones~\citep{google2025lens}. While fully local VLM inference avoids image transmission, deploying capable models on resource-constrained devices introduces memory and computational challenges~\citep{chu2024mobilevlm}. Even a 3B-parameter model requires roughly 6\,GB for 16-bit weights alone, before runtime overhead, although quantization can reduce this footprint. 
Model capacity can also matter: on the MMMU benchmark~\citep{yue2024mmmu}, GPT-4o achieves 70.3\%, outperforming Qwen2.5-VL-3B (53.1\%) by 17.2 percentage points~\citep{bai2025qwen25vltechnicalreport}. We therefore target settings where users choose cloud-hosted VLMs for answer generation while retaining local control over the visual information disclosed.

Existing approaches address related but distinct objectives. Query-agnostic anonymization suppresses predefined sensitive content regardless of whether it is needed for the task~\citep{yu2017iprivacy,orekondy2018connecting,sun2018natural,maximov2020ciagan}, while grounding-based methods identify query-relevant regions without determining which sensitive details within those regions are necessary~\citep{kamath2021mdetr,li2022grounded,liu2024grounding,zheng2024instruction}. 
These distinctions motivate a framework that considers query requirements, sensitive information within relevant regions, and potentially revealing contextual cues, while remaining independent of the downstream VLM.

To address this gap, we introduce \textbf{Query-Guided Visual Minimization (QGVM)}, a framework that determines what visual content to disclose before cloud inference. Given an image and a query, QGVM independently identifies query-relevant regions and detects potentially sensitive content, including faces, identifiers, and text. Crucially, identifying a relevant region does not imply that all information within it is necessary. 
QGVM therefore determines disclosure necessity at three levels: \emph{type}, whether a category of sensitive information is required; \emph{instance}, which detected item is relevant to the query; and \emph{content}, restricted to recognized text, which OCR strings should remain visible.
A frozen text-only language model assesses type and content necessity using the query and, when applicable, recognized text, while spatial grounding determines relevant instances. Finally, QGVM combines background suppression with layered redaction and restoration to preserve required visual details while concealing unnecessary information. 
The resulting minimized image is transmitted to a black-box cloud VLM, requiring no modifications to the downstream model.

We evaluate QGVM along two complementary privacy dimensions: \emph{explicit leakage}, measured by the recoverability of query-unnecessary sensitive information such as faces and text, and \emph{implicit leakage}, measured through geolocation attacks that exploit contextual scene cues. Our primary evaluation combines DoxBench~\citep{luo2025doxing} with Scene-Privacy, a curated, human-reviewed benchmark comprising 186 image--query cases over 65 images. We further assess generalization on independently held-out benchmarks and use a separate paired-query evaluation to test whether disclosure adapts to the information required by each query.


Our main contributions are:

\begin{itemize}

     
    \item \textbf{Necessity-aware visual minimization.}
    We introduce \textbf{QGVM}, which combines query grounding, sensitive-content detection, and type-, instance-, and content-level necessity decisions to protect unnecessary sensitive details even within query-relevant regions, independently of the downstream black-box VLM.


    \item \textbf{Human-reviewed evaluation resources for query-dependent privacy.}
    We extend DoxBench~\citep{luo2025doxing} with query-specific privacy annotations for 256 image--query cases across 149 images and introduce \textbf{Scene-Privacy}, comprising 186 cases across 65 images. Both sets undergo human review to support evaluation of query-necessary preservation and unnecessary disclosure. 
    We additionally introduce a reusable paired-query evaluation with 99 query links across 34 images to test whether disclosure adapts to changing query requirements.

    \item \textbf{Empirical findings and privacy--utility trade-offs.}
    Across multiple benchmarks, downstream VLMs, and privacy attackers, we demonstrate that identifying relevant regions alone is insufficient to prevent sensitive-information leakage. Comparisons with existing privacy methods, targeted ablations, and failure analysis establish the benefits of necessity-aware disclosure while revealing its limitations, particularly when answering queries requires preserving sensitive information.

\end{itemize}

With OpenAI GPT-6.1 Sol, QGVM achieves 85.3\% DoxBench answer correctness versus 77.5\% for PGP~\citep{azizian2026prompt}, while reducing Sol geolocation success within 1\,km from 25.7\% (unprotected) to 5.8\%. On Scene-Privacy, QGVM achieves 86.5\% correctness with Claude Opus-5.5 while reducing face re-identification from 27.7\% (PGP) to 4.5\%. On held-out BIV-Priv-Seg~\citep{tseng2025biv}, unnecessary private-text recovery falls from 33.9\% (PGP) to 15.3\%, with a 6.0-percentage-point reduction in judged answer preservation. Paired-query disclosure switching improves from 29.1\% with grounding-only cropping to 69.6\% with QGVM.

\section{Related Work}
\label{sec:related_work}

QGVM connects three lines of research: query-conditioned visual grounding, vision-based sensitive information leakage, and local--cloud data minimization controlling what is transmitted to remote models.

\subsection{Query-Conditioned Visual Grounding and Minimization}
Vision--language grounding methods such as KOSMOS-2~\citep{peng2024grounding} and Grounding DINO~\citep{liu2024grounding} localize regions described by text, while promptable segmentation methods refine spatial selections~\citep{kirillov2023segment,zhang2023faster}. Instruction-Guided Visual Masking (IVM)~\citep{zheng2024instruction} suppresses instruction-irrelevant regions, while Prompt-Guided Prefiltering (PGP)~\citep{azizian2026prompt} preserves query-relevant content to improve image compression for VLM inference. Related edge video analytics systems use task-aware filtering to reduce communication costs~\citep{li2020reducto} or optimize inference and retraining under resource constraints~\citep{bhardwaj2022ekya}. However, grounding a relevant region does not establish which sensitive details within it are necessary for the query.

\subsection{Visual Privacy and Image Anonymization}
Visual privacy research examines sensitive information explicitly visible in images and private attributes inferred from contextual cues~\citep{orekondy2017towards,tomekcce2024private,luo2025doxing}. Anonymization methods suppress or replace identifying content such as faces and bodies~\citep{maximov2020ciagan,hukkelaas2023deepprivacy2}, while recent approaches combine sensitive-content detection with context-aware anonymization and image editing~\citep{dinh2026unsafe2safe,aufschlager2026towards}. Other directions include privacy-aware model adaptation~\citep{samson2024little} and instruction-guided image editing~\citep{brooks2023instructpix2pix,zhang2023magicbrush}.

\subsection{Data Minimization in Local--Cloud Systems}
Prior work investigates text anonymization and privacy leakage in language-model interactions~\citep{pilan2022text,dou2024reducing,staab2024beyond,staab2024large}. AirGapAgent~\citep{bagdasarian2024airgapagent} restricts access to task-relevant private information, while PAPILLON~\citep{siyan2025papillon} combines local and remote language models to reduce sensitive-data disclosure. Contextual-integrity evaluations further study context-dependent privacy expectations~\citep{mireshghallah2024can,shao2024privacylens}. Recent multimodal work includes MPCI-Bench~\citep{wang2026mpci}, granular location-disclosure controls~\citep{mendes2024granular}, and local visual instruction rewriting that converts multimodal requests into text-only representations~\citep{mishra2025revision}. Unlike text-only rewriting, QGVM retains the image modality while minimizing unnecessary visual disclosure before cloud inference.

\begin{figure*}[t]
    \centering
    \includegraphics[width=\textwidth]{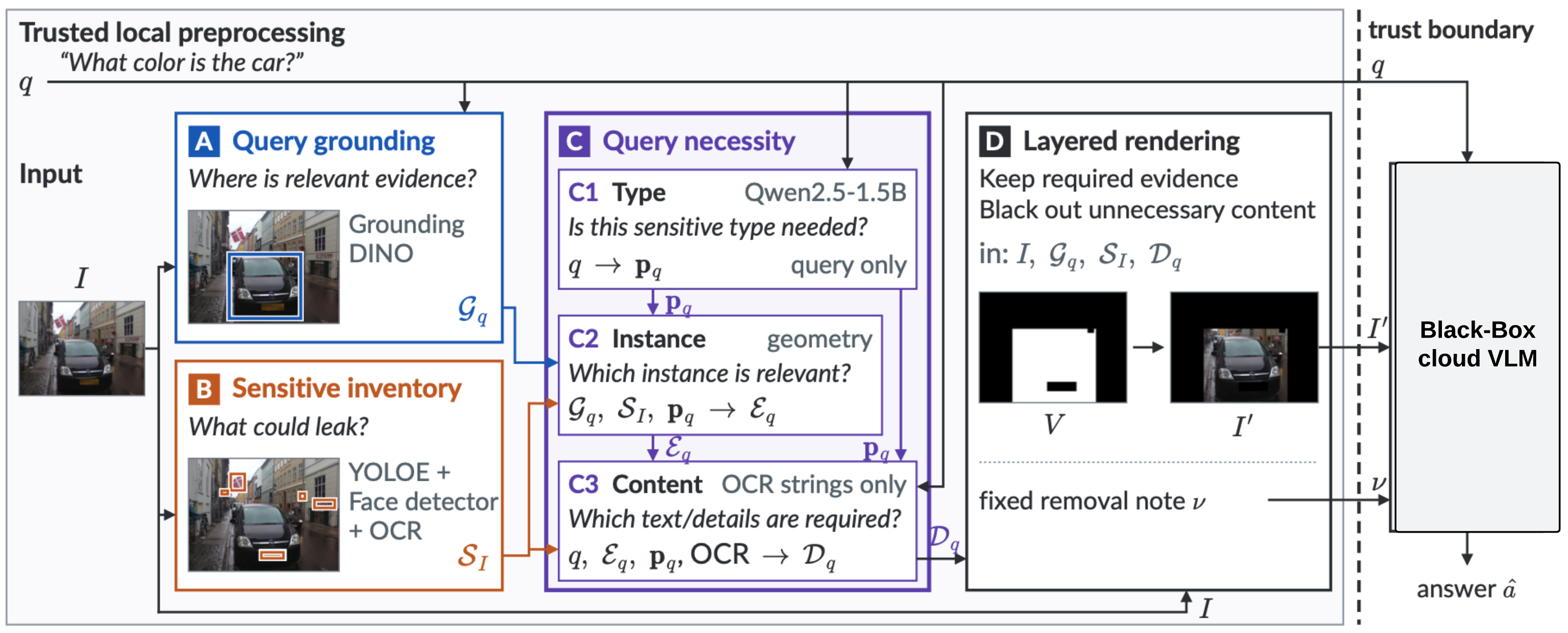}
    \caption{\textbf{Overview of QGVM.} Query grounding (A) localizes task-relevant content, while an independent inventory (B) detects potentially sensitive content. Query necessity (C) determines which sensitive types, instances, and text details are required. Layered rendering (D) produces the minimized image. Only the minimized image, query, and removal note cross the cloud trust boundary. Plate characters are obscured solely for publication.}
    \label{fig:qgvm}
\end{figure*}

\section{QGVM: Query-Guided Visual Minimization}
\label{sec:method}

\subsection{Problem and Design Principle}
\label{sec:problem}

Consider an image $I$ and a natural-language query $q$ intended for a cloud vision-language model $\mathcal{C}$. A trusted local minimizer $\mathcal{M}$ produces $I'=\mathcal{M}(I,q)$ before transmission. The cloud receives only the transformed image, the query, and a short note indicating that some visual regions were intentionally removed; it never receives $I$.

Our guiding principle is to reveal \emph{what the query needs}, not everything in the region to which it refers. A query about a vehicle's color may require its body but not its license plate; a query asking for that plate must instead retain its text. We distinguish \emph{relevance} (where content for the query is located) from \emph{necessity} (whether a particular sensitive type, instance, or content unit must be visible to answer). The distinction matters because private information can be directly readable or inferred from seemingly innocuous context. QGVM approximates minimal sufficient disclosure through fixed, query-conditioned decisions. It does not claim to find a mathematically minimal image or to guarantee privacy against arbitrary inference.

The output is a pixel-level disclosure mask $V(I,q)\in\{0,1\}^{H\times W}$ applied to the original image. For each pixel location $x=(h,w)$, the minimized image is
\begin{equation}
    I'(x)=V(x;I,q)\,I(x),
    \label{eq:qgvm_render}
\end{equation}
So each transmitted pixel is either unchanged or black. Figure~\ref{fig:qgvm} illustrates the four-stage pipeline described below.

\subsection{Grounding and Sensitive-Item Inventory}
\label{sec:inventory}

\paragraph{A. Grounding (where to look).}
A frozen Grounding DINO-base detector takes the query as its text prompt and returns a set of query-relevant boxes $\mathcal{G}(I,q)$. We remove generic detections and boxes covering more than half the image, retaining at most ten targets. Grounding supplies spatial support for the query but does not decide whether sensitive attributes inside the target are needed. For instance, grounding a car does not justify disclosing its plate. Additional implementation details, including detector thresholds, are provided in the supplementary material.

\paragraph{B. Inventory: what could leak.}
Independently of grounding, we construct a \emph{flat} inventory $\mathcal{S}(I)=\{(B_i,t_i,z_i)\}$, where $B_i$ localizes an item, $t_i$ denotes its sensitive category, and $z_i$ contains recognized text when available. YOLOE-v8l-seg~\citep{wang2025yoloerealtimeseeing} proposes categories such as license plates, documents, screens, tattoos, signs, and text-bearing objects. InsightFace~\citep{guo2022scrfd} detects faces on the full image and person crops, followed by a tiled zoom audit for small faces. EasyOCR~\citep{jaidedai2020easyocr} extracts text boxes and strings from overlapping image tiles; additional text-carrier detections protect signs or documents whose text OCR misses.

\subsection{Three Levels of Query Necessity}
\label{sec:necessity}

\paragraph{C1. Type necessity.}
We use a frozen Qwen2.5-1.5B-Instruct~\citep{qwen2025qwen25technicalreport} language model as a compact \emph{text-only} necessity estimator. We consider eight predefined sensitive types aligned with our detector inventory: face, license plate, text, screen, document, tattoo, medical information (e.g., medication labels), and location. For each type $t$, it receives the query and asks whether seeing that type of information is necessary to answer. We normalize the next-token logits for \texttt{Yes} and \texttt{No}:
\begin{equation}
    p_t(q)=\frac{\exp(\ell_t^{\mathrm{Yes}})}
    {\exp(\ell_t^{\mathrm{Yes}})+\exp(\ell_t^{\mathrm{No}})}
    \label{eq:type_necessity}
\end{equation}
We mark type $t$ as needed when $p_t(q)>\tau_t$, using $\tau_t=0.2$ for ordinary sensitive types and the more permissive $\tau_{\mathrm{text}}=0.03$ for text. These values are fixed development-set operating thresholds; the normalized logits are decision scores, not calibrated probabilities of true necessity. The low text threshold favors recall for reading questions; content-level selection below narrows uncertain disclosures. Importantly, \emph{this type decision sees the query only}: neither image pixels nor the downstream model's answer are inputs.

\paragraph{C2. Instance necessity.}
A needed type does not imply that every instance should be disclosed. If the grounding stage found targets, a sensitive item is eligible only when at least half its area lies inside a grounded box:
\begin{equation}
    g_i(I,q)=\mathbf{1}\!\left[
    \max_{G\in\mathcal{G}(I,q)}
    \frac{|B_i\cap G|}{|B_i|}\geq 0.5\right]
    \label{eq:instance_gate}
\end{equation}
If grounding returns no target, all instances of a needed type remain eligible. To reduce false exclusions, when a required type has no eligible item, we retry against targets expanded by 50\% and otherwise select the nearest item. This allows a query referring to one person or object to preserve the relevant instance without automatically revealing every other face, plate, or text region in the image.

\paragraph{C3. Content necessity.}
Text often requires finer control than an object or text box. We therefore use \emph{confidence-adaptive content disclosure}. When $p_{\mathrm{text}}(q)>0.2$, QGVM preserves text within the selected relevant region, favoring completeness for explicit reading tasks. In the uncertain range $0.03<p_{\mathrm{text}}(q)\leq0.2$, the same text-only model instead evaluates each recognized OCR string against the query and discloses only strings scoring above $0.2$, falling back to the highest-scoring string if none qualify. Below $0.03$, text is not selected on type necessity alone. Thus, the type scorer uses \emph{only the query}, whereas the content scorer additionally reads OCR strings, but neither processes raw image pixels. Embedded OCR text inherits disclosure from an explicitly required carrier such as a document, screen, or plate, whereas a face does not inherit disclosure merely because its containing person is relevant. These containment rules avoid hiding content needed to read a selected object while protecting unrelated identifiers.

\subsection{Background Suppression and Layered Rendering}
\label{sec:rendering}

\paragraph{D. Composing the disclosure mask.}
For queries referring to localized objects, we suppress contextual background when at least one target is grounded. The initial visible area is the union of target boxes expanded by 10\% (at least 16 pixels) and individually disclosed sensitive items. Whole-scene queries, identified by a fixed set of contextual terms, bypass this background-suppression rule so that potentially necessary scene cues remain available. The inventory then determines which sensitive items to hide or restore inside the visible area.

To compose overlapping decisions, QGVM first blacks out the suppressed background, then paints sensitive items from largest to smallest. Suppressed items are blacked out with an additional 15\% margin; disclosed items restore their original pixels. Consequently, a relevant object can remain visible while its plate or face is removed, and a requested text detail can be restored within an otherwise suppressed carrier. Finer regions take precedence, and the final image obeys Eq.~\ref{eq:qgvm_render}. The cloud receives $I'$ together with a fixed note that black regions were removed intentionally and that the answer should rely on visible content. The note is part of the common inference interface, not evidence of privacy by itself.

\noindent \textbf{Training-free implementation and scope.}
All perception modules and the necessity model are frozen; QGVM uses no query-specific training or cloud-model feedback.
For memory efficiency, the necessity model uses unquantized BF16 weights, with decoder layers stored in pinned CPU memory and transferred sequentially to the GPU for inference. Grounding and OCR detection run in half precision. We evaluate recoverable leakage empirically rather than claiming a formal privacy guarantee.

\section{Experiments}
\label{sec:experiments}

We evaluate whether query-necessity estimation enables more selective visual disclosure than relevance-based filtering alone. In particular, we investigate four questions: \textbf{(Q1)} Can QGVM reduce unnecessary privacy exposure while preserving utility? \textbf{(Q2)} Does disclosure adapt to whether sensitive information is required by the query?
\textbf{(Q3)} How do individual QGVM components affect the privacy--utility trade-off? \textbf{(Q4)} Do these findings extend across visual domains, downstream models, and privacy attackers?

\begin{table*}[t]
\centering
\small
\setlength{\tabcolsep}{3pt}
\renewcommand{\arraystretch}{1.05}
\caption{\textbf{Primary-benchmark privacy--utility results} (\%).
Qw: Qwen2.5-VL-7B correctness (original resolution);
Opus: Claude Opus~5.5;
Sol: OpenAI Sol.
Correctness is measured using the indicated answerer.
Op@25 and Sol@1/@25 denote geolocation attack success
within 25 and 1/25\,km, respectively.
Face and Text denote Scene-Privacy leakage.
Cloud correctness and all privacy attacks use
transmitted-resolution images.
Higher correctness and lower leakage are preferred.
Among privacy-preserving methods, best results are shown in \textbf{bold} and second-best results are \underline{underlined}.
}
\label{tab:main_privacy}

\begin{tabular}{@{}l*{11}{r}@{}}
\toprule
& \multicolumn{6}{c}{\textbf{DoxBench}}
& \multicolumn{5}{c}{\textbf{Scene-Privacy}} \\
\cmidrule(lr){2-7}\cmidrule(l){8-12}

& \multicolumn{3}{c}{Correctness $\uparrow$}
& \multicolumn{3}{c}{Geo success $\downarrow$}
& \multicolumn{3}{c}{Correctness $\uparrow$}
& \multicolumn{2}{c}{Leakage $\downarrow$} \\
\cmidrule(lr){2-4}
\cmidrule(lr){5-7}
\cmidrule(lr){8-10}
\cmidrule(l){11-12}

Method
& Qw & Opus & Sol
& Op@25 & Sol@1 & Sol@25
& Qw & Opus & Sol
& Face & Text \\
\midrule

No protection
& 63.4 & 82.2 & 82.6
& 58.3 & 25.7 & 35.3
& 64.0 & 91.1 & 86.7
& 98.9 & 80.2 \\

Query-agnostic filter
& 17.6 & 40.6 & 41.7
& \underline{23.0} & 6.7 & 15.8
& 20.4 & 28.9 & 30.4
& \textbf{0.0} & \textbf{10.0} \\

IVM
& 21.2 & 46.6 & 49.3
& \textbf{14.3} & \textbf{3.1} & \textbf{4.0}
& 26.9 & 41.7 & 49.1
& 17.0 & \underline{17.9} \\

KOSMOS-2
& 67.1 & \underline{78.8} & 79.7
& 53.0 & 20.6 & 29.3
& 54.3 & 71.4 & 70.4
& 73.1 & 58.8 \\

Crop-to-target
& \underline{67.8} & 78.6 & \underline{83.5}
& 32.6 & 9.8 & 18.9
& 64.2 & \underline{86.0} & 86.1
& 57.4 & 45.8 \\

PGP
& 65.2 & \underline{78.8} & 77.5
& 36.0 & 10.5 & 16.8
& \underline{64.8} & 83.8 & \underline{86.3}
& 27.7 & 29.0 \\

VLM-as-redactor
& 58.2 & 68.3 & 68.4
& 29.2 & 9.6 & 18.6
& 64.7 & 84.8 & 82.7
& 47.1 & 32.5 \\

\midrule
\textbf{QGVM (ours)}
& \textbf{67.9} & \textbf{79.0} & \textbf{85.3}
& 34.2 & \underline{5.8} & \underline{15.4}
& \textbf{64.9} & \textbf{86.5} & \textbf{86.5}
& \underline{4.5} & 23.3 \\

\bottomrule
\end{tabular}
\end{table*}

\subsection{Experimental Setup}
\label{sec:exp_setup}

\paragraph{Benchmarks.}
Our primary evaluation focuses on two human-reviewed privacy benchmarks: \textbf{DoxBench}, containing 256 image--query cases over 149 images, and \textbf{Scene-Privacy}, containing 186 cases over 65 images. Both evaluate visual question answering in the presence of potentially unnecessary sensitive information. DoxBench additionally supports inferential geolocation attacks, while Scene-Privacy enables direct evaluation of face re-identification and private-text recovery. 

To assess generalization beyond the primary benchmarks, we evaluate five additional datasets. BIV-Priv-Seg~\citep{tseng2025biv} provides 651 held-out images and 1,071 queries that either require (Track S) or do not require (Track N) private text. OSV-5M~\citep{astruc2024openstreetview} provides 799 street-scene images for measuring incidental geographic disclosure. PrivScreen-450~\citep{zhang2026dualtap} evaluates screenshot understanding and sensitive-information recovery across 450 screenshots from ten applications. RefCOCOg~\citep{mao2016generation} (500 examples) and PointArena~\citep{cheng2025pointarena} (829 examples) measure spatial-grounding utility. We further construct a paired-query evaluation comprising 99 links across 34 images to measure query-dependent disclosure directly. Together, these datasets evaluate the QGVM framework across diverse visual domains and query requirements, with BIV-Priv-Seg providing the primary held-out test. Dataset provenance and split details are documented in the supplementary material. \\

\noindent \textbf{Baselines.}
We compare against unprotected images, query-agnostic sensitive-item removal, IVM~\citep{zheng2024instruction}, KOSMOS-2~\citep{peng2024grounding}, and PGP~\citep{azizian2026prompt}. To isolate the contribution of necessity estimation, we include two additional baselines. \emph{Crop-to-target} uses QGVM's own query grounding but omits sensitive-item detection and necessity estimation, separating localization from selective disclosure. \emph{VLM-as-redactor} prompts a frozen Qwen2.5-VL-7B model with the image and query to identify regions to preserve or redact, testing whether direct multimodal reasoning can replace explicit necessity estimation. Its prompt is selected using development data and frozen before evaluation. All methods use the same black-redaction operator. \\

\noindent \textbf{Evaluation protocol.}
We evaluate three frozen local VLMs: Qwen2.5-VL-7B, Qwen3-VL-8B, and InternVL3.5-8B. For cloud evaluation, we use Claude Opus~5.5~\citep{anthropic2026opus55} and OpenAI GPT-6.1 Sol on both primary benchmarks, covering all eight methods and 442 image--query cases. We use the same fixed removal note for all transformed images and evaluate privacy leakage on the exact resized images transmitted to the cloud. On DoxBench and Scene-Privacy, a fixed grader assesses answer correctness against reference answers. On OSV-5M and BIV-Priv-Seg, we measure \emph{judged answer preservation}, assessing whether a transformed-image response retains the key information in the same model's original-image response, rather than independently verified correctness. Only cases with informative original-image responses contribute to this metric, as reflected in the metric-specific sample sizes in Table~\ref{tab:generalization}.

Privacy is evaluated through attacks on the disclosed images. We measure geographic leakage using GeoCLIP~\citep{vivanco2023geoclip} and VLM-based geolocation attackers, face re-identification using ArcFace~\citep{deng2019arcface} at a 0.1\% false-accept threshold, and private-text recovery using VLM transcribers relative to their original-image transcriptions. BIV-Priv-Seg additionally provides ground-truth masks for measuring private-pixel exposure. Every transformed-image prompt includes the same note indicating that black regions were intentionally removed; original images receive no note. We report source-image-balanced results with paired cluster-bootstrap confidence intervals (10,000 resamples) and Holm correction within predefined comparison families; PrivScreen is clustered by application. Further implementation details and statistical protocols are provided in the supplementary material.

\begin{figure*}[!h]
    \centering
    \includegraphics[width=\textwidth]{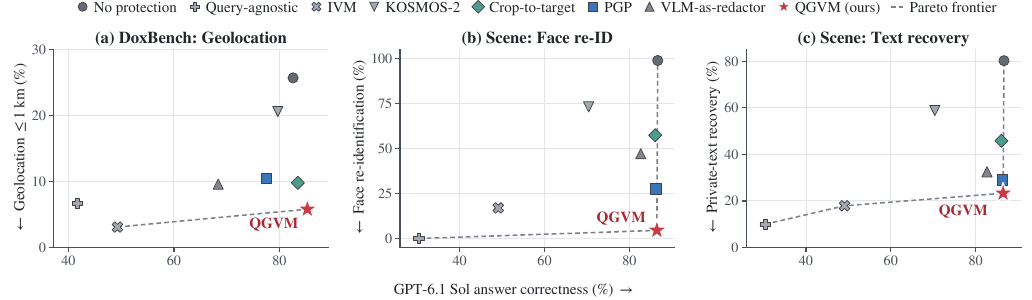}
    \caption{\textbf{Privacy--utility trade-offs on primary benchmarks.}
    GPT-6.1 Sol answer correctness ($x$, higher is better) versus privacy attack success ($y$, lower is better).
    Panels show (a) DoxBench geolocation within 1\,km using a separately prompted Sol attacker, (b) Scene-Privacy face re-identification, and (c) private-text recovery.
    Dashed lines indicate point-estimate Pareto frontiers; QGVM lies on all three.}
    \label{fig:privacy_utility}
\end{figure*}

\subsection{Q1: Privacy--Utility Trade-off}
\label{sec:exp_tradeoff}

\noindent \textbf{Preserving utility across downstream VLMs.}
Table~\ref{tab:main_privacy} reports correctness on DoxBench and Scene-Privacy, respectively. QGVM achieves 67.9\% and 64.9\% with Qwen2.5-VL-7B (vs.\ 63.4\% and 64.0\% unprotected), 79.0\% and 86.5\% with cloud Opus~5.5 (vs.\ 82.2\% and 91.1\%), and 85.3\% and 86.5\% with GPT-6.1 Sol (vs.\ 82.6\% and 86.7\%). QGVM achieves the highest observed correctness among protection methods in all six settings. Notably, on DoxBench with Sol, QGVM significantly outperforms PGP (77.5\%) by 7.8 percentage points (Holm-adjusted $p=0.007$).

\noindent \textbf{Protecting sensitive details beyond grounding.}
On Scene-Privacy, QGVM reduces face re-identification to 4.5\%, compared with 27.7\% for PGP, 57.4\% for Crop-to-target, and 47.1\% for VLM-as-redactor, and reductions are significant after Holm correction. QGVM also reduces private-text recovery to 23.3\%, compared with 45.8\% for cropping and 29.0\% for PGP.
QGVM maintains 86.5\% Scene-Privacy correctness with both Opus and Sol despite substantially lower face and text leakage. Thus, grounding alone does not protect unnecessary sensitive details within relevant regions.

\noindent \textbf{Geolocation leakage across attackers.}
On DoxBench, QGVM reduces Opus~5.5 geolocation success within 25\,km from 58.3\% (unprotected) to 34.2\%. With Sol as the attacker, success falls from 35.3\% to 15.4\% at 25\,km and from 25.7\% to 5.8\% at 1\,km. At 1\,km, QGVM also significantly improves over cropping (5.8\% vs.\ 9.8\%; Holm-adjusted). Jointly, QGVM answers correctly without successful Sol localization within 25\,km in 71.6\% of cases, significantly above the 53.7\% unprotected rate. However, Opus geolocation differences against PGP and cropping are not significant; more aggressive filtering achieves lower leakage at substantial utility cost.

Figure~\ref{fig:privacy_utility} visualizes these trade-offs. QGVM lies on the observed privacy--utility Pareto frontier across all three metrics, highlighting strong identity and text protection at competitive utility preservation.

\subsection{Q2: Query-Dependent Disclosure}
\label{sec:exp_query_dependence}

We directly evaluate whether disclosure adapts to query necessity using paired questions about the same sensitive target in a fixed image. One question requires the target information, whereas the other does not. We measure \emph{required retention} (RR), the fraction of needed targets that remain recoverable; \emph{unnecessary suppression} (US), the fraction of unnecessary targets that are not recoverable; and the \emph{correct disclosure switch rate} (CDSR), the fraction of paired queries satisfying both conditions.

\begin{table}[!h]
\centering
\small
\setlength{\tabcolsep}{5pt}
\caption{\textbf{Query-dependent disclosure} on 99 paired-query links across 34 images, evaluated using Qwen3-VL. CDSR measures joint retention when needed and suppression otherwise. 
}
\label{tab:query_dependence}
\begin{tabular}{lrrr}
\toprule
Method & RR $\uparrow$ & US $\uparrow$ & CDSR $\uparrow$ \\
\midrule
No protection         & 100.0 &  0.0 &  0.0 \\
Query-agnostic filter &   9.8 & \textbf{90.2} &  0.0 \\
IVM                   &  36.0 & 65.0 & 10.9 \\
KOSMOS-2              &  70.1 & 24.5 & 17.6 \\
Crop-to-target        &  \underline{94.1} & 34.3 & 29.1 \\
PGP                   &  90.7 & 60.6 & \underline{51.3} \\
VLM-as-redactor       & 88.2 & 56.1 & 47.7 \\
\midrule
\textbf{QGVM}         & \textbf{97.1} & \underline{71.8} & \textbf{69.6} \\
\bottomrule
\end{tabular}
\end{table}

As shown in Table~\ref{tab:query_dependence}, QGVM achieves 69.6\% CDSR, compared with 51.3\% for PGP, 47.7\% for VLM-as-redactor, and 29.1\% for Crop-to-target. The improvement over Crop-to-target is statistically significant after Holm correction, providing direct evidence that query grounding alone is insufficient for selective disclosure. Relative to PGP, QGVM improves CDSR by 18.3 percentage points. Together, these results demonstrate that explicit necessity estimation enables more reliable query-dependent disclosure than relevance-only selection, with promising improvements over the baselines.

\begin{table}[!h]
\centering
\small
\setlength{\tabcolsep}{4pt}
\caption{\textbf{Component ablations on development data.} Changes relative to full QGVM in percentage points. BIV-S measures reading utility ($\uparrow$), BIV-N measures private-text recovery ($\downarrow$), and OSV measures GeoCLIP@750\,km ($\downarrow$). Full QGVM scores are 56.0, 13.5, and 30.0, respectively. $^\dagger$ indicates an unadjusted 95\% confidence interval excluding zero.}
\label{tab:ablation}
\begin{tabular}{lrrr}
\toprule
Variant & BIV-S $\uparrow$ & BIV-N $\downarrow$ & OSV $\downarrow$ \\
\midrule
Keyword necessity        & $-3.6$ & $-5.6^\dagger$ & $+0.0$ \\
Disclose all text        & $+11.9^\dagger$ & $+8.7^\dagger$ & $+0.0$ \\
Always strict selection  & $-17.9^\dagger$ & $-4.8^\dagger$ & $+0.0$ \\
Without layered rendering & $-22.6^\dagger$ & $-1.6$ & $+0.0$ \\
Without background removal & $+0.0$ & $+0.8$ & $+15.5^\dagger$ \\
\bottomrule
\end{tabular}
\end{table}


\subsection{Q3: Component Analysis}
\label{sec:exp_ablation}

We isolate the contributions of necessity estimation, content-level selection, layered rendering, and background suppression through controlled ablations.

\noindent \textbf{The role of necessity estimation.}
Replacing the query-only necessity estimator with keyword matching reduces Scene-Privacy correctness by 7.1 percentage points, and BIV Track N judged preservation by 4.4 points, while also suppressing more private text. This highlights the role of explicit necessity estimation in preserving useful content rather than indiscriminately removing sensitive content. Additional necessity-prediction diagnostics are provided in the supplementary material.

\noindent \textbf{Granularity controls the privacy--utility trade-off.}
Table~\ref{tab:ablation} shows that disclosing all text within needed regions improves BIV reading utility by 11.9 points but increases unnecessary text recovery by 8.7 points. Conversely, always enforcing strict string selection reduces text leakage but costs 17.9 points in reading utility. Removing layered rendering further decreases reading utility by 22.6 points. These results motivate confidence-adaptive disclosure and layered restoration to preserve query-required details while limiting unnecessary exposure.

\noindent \textbf{Separating contextual from item-level privacy.}
Removing background suppression increases OSV GeoCLIP@750\,km leakage by 15.5 points and DoxBench Qwen3-VL Geo@25\,km leakage by 9.1 points. Together with the grounding-only control in Section~\ref{sec:exp_tradeoff}, these results distinguish two complementary mechanisms: background suppression limits contextual geographic content, whereas necessity-aware selection controls sensitive information within task-relevant regions.

\noindent \textbf{Staged vs.\ flat necessity estimation.}
We compare QGVM with Flat Item Necessity (FIN), which replaces staged type--instance--content decisions with a single LM decision per detected item, using the same grounding, inventory, and rendering. FIN's threshold is selected on development data to match or improve text privacy. On held-out BIV-Priv-Seg, QGVM achieves substantially higher judged preservation for private-text reading (56.8\% vs.\ 32.9\%). On Scene-Privacy, QGVM reduces face re-identification (4.5\% vs.\ 21.5\%) and improves Sol answer correctness (86.5\% vs.\ 77.6\%). These findings favor the complete staged design.

\begin{table}[!t]
\centering
\small
\setlength{\tabcolsep}{6pt}
\caption{\textbf{Generalization beyond the primary benchmarks.} BIV is held out. All values are percentages; $n$ denotes the number of contributing evaluation cases for each metric. P denotes judged answer preservation,
which includes only informative original-image responses.
Higher utility and lower leakage are preferred.}
\label{tab:generalization}
\begin{tabular}{@{}llrrrrl@{}}
\toprule
Bench & Metric & $n$ & Crop & PGP & QGVM  \\
\midrule
BIV-N & Private text recovered $\downarrow$ & 373 & 74.9 & 33.9 & 15.3 \\
BIV-N & Answer preserved (P) $\uparrow$ & 501 & 82.4 & 62.7 & 56.7  \\
\midrule
OSV & GeoCLIP $\leq 750\,\mathrm{km}$ $\downarrow$ & 799 & 25.0 & 30.3 & 25.7  \\
OSV & Answer preserved (P) $\uparrow$ & 260 & 85.0 & 84.2 & 83.1  \\
\bottomrule
\end{tabular}
\end{table}

\subsection{Q4: Generalization and Practicality}
\label{sec:exp_generalization}

\paragraph{Generalization beyond the design sets.}
Table~\ref{tab:generalization} summarizes evaluations beyond the primary benchmarks. On held-out BIV-Priv-Seg, QGVM reduces query-unnecessary private-text recovery to 15.3\%, compared with 33.9\% for PGP, 74.9\% for crop-to-target, and 62.3\% for VLM-as-redactor. The reductions are statistically significant, and a second text transcriber confirms the trend. However, this privacy advantage reduces judged answer preservation by 6.0 percentage points relative to PGP on BIV-N. Additional privacy--utility results for both BIV tracks are reported in the supplement. On OSV-5M, QGVM reduces GeoCLIP geolocation success within 750\,km from 30.3\% to 25.7\% relative to PGP, without a statistically significant difference in judged answer preservation. However, cropping provides comparable location protection; the improvement is attacker-dependent. Overall, QGVM offers stronger protection against unnecessary private-text disclosure, but does not uniformly improve the privacy--utility trade-off.


\noindent \textbf{Evaluator and interface robustness.}
We assess sensitivity to downstream VLMs and privacy attackers. On BIV, QGVM's private-text protection persists under a second VLM transcriber (16.8\% recovery versus 35.5\% for PGP), whereas relative geolocation performance varies across attackers. The fixed removal note also affects downstream utility: omitting it reduces OSV development-set judged answer preservation from 82.6\% to 67.4\%. We therefore apply the same note to all transformed-image methods and report its effect separately.


\noindent \textbf{Failure analysis.}
We analyze 218 traced failures to identify limitations of selective disclosure. Necessity-estimation errors account for 68\% of examined BIV reading failures, primarily through overly strict text selection or missed required types. Conversely, 28 of 30 examined query-unnecessary private-text leaks result from necessity false positives. These findings reveal two failure modes: suppressing information needed to answer the query and retaining sensitive information that should have been removed.

\noindent \textbf{System profiling.}
We profile QGVM on 100 image--query cases using an NVIDIA A10 GPU. Steady-state local preprocessing takes
1.11\,s at the median and 1.74\,s at the 90th percentile, with 2.80\,GiB peak process-level GPU memory. Sensitive-item inventory dominates runtime, primarily due to tiled OCR and face auditing. On the same GPU, local Qwen2.5-VL-7B inference requires 16.0\,GiB but has a lower median latency (0.75\,s). Since these systems perform different tasks, this comparison establishes QGVM's smaller measured GPU footprint, not an end-to-end speed or mobile-deployment advantage.

\section{Conclusion}
\label{sec:conclusion}

We introduced QGVM, a framework that separates query relevance from disclosure necessity for cloud VLM inference. Across multiple VLMs and privacy attackers, QGVM reduces sensitive-information leakage beyond relevance-only selection while preserving competitive answer correctness. With GPT-6.1 Sol, QGVM achieves 85.3\% DoxBench correctness versus 77.5\% for PGP. Held-out evaluations and a matched flat-necessity baseline further support the staged design, although necessity errors can impair text-reading utility and geolocation protection remains attacker-dependent.





{
    \small
    \bibliographystyle{ieeenat_fullname}
    \bibliography{main}
}

\maketitlesupplementary


\appendix

\section{Implementation of QGVM}\label{sec:impl}
This supplement details the four-stage implementation used for the experiments in the main paper. We denote the canonical EXIF-transposed input by $I\in[0,255]^{H\times W\times3}$ and the query by $q$. The final transformed image has the form $I'=V(I,q)\odot I$, with $V\in\{0,1\}^{H\times W}$; therefore, the locally produced image consists only of unchanged pixels and black pixels (before the resizing shared by all methods for cloud transmission). No diffusion, image synthesis, or pixel substitution is used.

\paragraph{1. Grounding.} A frozen Grounding DINO-base checkpoint receives the normalized query as its text prompt. The normalization removes a leading ``point to/at/out,'' lowercases text, strips trailing punctuation, and adds a period. We use box and text thresholds of 0.30 and 0.25, discard boxes whose labels contain no nongeneric word or occupy more than half the image, and keep at most ten targets. No additional query rewriting is performed. If grounding produces no target, contextual background suppression is disabled, but independently detected sensitive items are still processed.

\paragraph{2. Sensitive-item inventory.} Independently of grounding, we construct a flat inventory of faces, license plates, text, screens, documents, tattoos, medical labels, and location-bearing signs. YOLOE-v8l-seg proposes the object classes (confidence 0.25, IoU 0.7, maximum 100 detections). Face detection uses InsightFace \texttt{buffalo\_l} (SCRFD; detection threshold 0.5 at $640\times640$) on the full image and person crops, followed by an audit on a fixed overlapping $3\times3$ tile grid and low-threshold person proposals. EasyOCR 1.7.2 runs on overlapping tiles: the image long side is capped at 2560 pixels, each tile at 1280 pixels, and tile overlap is 64 pixels. Separate low-threshold YOLOE carriers catch text-bearing objects missed by OCR. Contained duplicates are suppressed using the inventory-specific 50\% overlap conditions. Unobserved items cannot be guaranteed protection.

\paragraph{3. Three necessity levels.} For each of eight sensitive types $t$, a frozen \texttt{Qwen2.5-1.5B-Instruct} model receives only the text query and a type descriptor. Its next-token Yes/No logits define a \emph{decision score}, not a calibrated probability:
\begin{equation}\label{eq:necessity}
 p_t(q)=\frac{e^{\ell_{t,\mathrm{Yes}}}}{e^{\ell_{t,\mathrm{Yes}}}+e^{\ell_{t,\mathrm{No}}}}.
\end{equation}
The type is needed if $p_t(q)>0.2$ except for \texttt{text}, which uses $0.03$. For a needed type, an item is disclosed when at least half its area lies inside any grounded target; if no target is grounded, all items of a needed type are disclosed. If targets exist but no item of a needed type is disclosed, the method discloses the items of that type that overlap a target expanded by 50\% per side, and otherwise the single item nearest to a target.

The text decision uses two regimes. For $p_{\rm text}>0.2$ (generous regime), OCR-string selection is off and needed text inside disclosed carriers is retained. For $p_{\rm text}\leq0.2$ (strict regime), each recognized OCR string that is already disclosed by the type and instance rules, or that lies at least half inside a disclosed non-text carrier (e.g., a needed document, screen, or plate), is scored separately by a query-and-OCR-string prompt; strings with scores above 0.2 are disclosed, with a highest-scoring fallback if none passes. Because text is a needed type only when $p_{\rm text}>0.03$, for $p_{\rm text}\leq0.03$ this scoring reaches only strings inside disclosed non-text carriers. Unread text boxes inside disclosed carriers inherit disclosure in both regimes, but a face never inherits disclosure merely because its person or carrier is relevant. Thus type scoring is query-only, while OCR content scoring is conditional on recognized strings; neither stage reads raw image pixels directly.

\paragraph{4. Background suppression and rendering.} A fixed set of contextual words (including \emph{scene, location, background, room, sky, nearby, street}) exempts whole-scene queries from background suppression. Otherwise, if targets exist, the visible context is the union of grounded boxes and disclosed-item boxes, each extended on every side by the larger of 16 pixels and 10\% of the box dimension. Sensitive items are then processed from larger to smaller bounding boxes. A suppressed item is blacked out with a per-side margin of $\lfloor0.15d\rfloor+2$ pixels; disclosed items restore their \emph{exact} boxes from the original image. The painter order lets small retained text appear within an otherwise suppressed carrier and prevents a disclosed region from automatically exposing a small sensitive face. The output is converted into the same binary keep-mask and black renderer used for all comparison methods.

\section{Exact Prompt Interface}\label{sec:prompts}
Table~\ref{tab:types} lists the exact descriptor strings for the eight type-level queries. The method uses the following frozen text, with the whitespace-stripped query substituted for \texttt{\{q\}} and the descriptor for \texttt{\{d\}}; each prompt is wrapped in the model's chat template before scoring.

\begin{promptpanel}{Stage C \,$\cdot$\, Type-necessity prompt (one prompt per sensitive type, scored in a single batch)}
\begin{chatmsg}{System}{sysrole}
You judge what an assistant must see in a photo to answer a question. Answer Yes only if the question cannot be answered without it.
\end{chatmsg}
\begin{chatmsg}{User}{userrole}
Question about a photo: "@{q}@"
To answer this question, must the assistant see @{d}@? Answer Yes or No.
\end{chatmsg}
\tcblower
\slotname{q}~user query \quad
\slotname{d}~type descriptor (Table~\ref{tab:types})\\[1pt]
\textbf{Score:} $p_t=\mathrm{softmax}(\ell_{\mathrm{Yes}},\ell_{\mathrm{No}})_{\mathrm{Yes}}$ at the next token;
type needed if $p_t>\tau_t$ ($\tau=0.2$, $\tau_{\text{text}}=0.03$).
\end{promptpanel}

For strict text selection, the same model (with the same system prompt) instead receives, with the OCR string truncated to 120 characters:

\begin{promptpanel}{Stage C \,$\cdot$\, OCR-string prompt (strict regime, $p_{\text{text}}\leq0.2$)}
\begin{chatmsg}{System}{sysrole}
You judge what an assistant must see in a photo to answer a question. Answer Yes only if the question cannot be answered without it.
\end{chatmsg}
\begin{chatmsg}{User}{userrole}
Question about a photo: "@{q}@"
A piece of text in the photo reads: "@{s}@"
Is reading this piece of text needed to answer the question? Answer Yes or No.
\end{chatmsg}
\tcblower
\slotname{s}~OCR string read locally (truncated to 120 characters). Disclosed if $p>0.2$; if none passes, the top-scoring string is kept.
\end{promptpanel}

\begin{table}[t]
\centering\footnotesize
\caption{Verbatim type descriptors used in necessity prompting.}
\label{tab:types}
\begin{tabularx}{\columnwidth}{@{}lX@{}}\toprule
Type & Descriptor substituted for \texttt{\{d\}} \\\midrule
Face & a person's face (for identity, expression, gaze, glasses, makeup, facial hair or age)\\
Plate & a vehicle's license plate\\
Text & written text, numbers, names, logos or brand names that must be read\\
Screen & the content shown on a screen\\
Document & the content of a document, card, ticket, ID or receipt\\
Tattoo & a tattoo\\
Medical & a medicine or prescription label\\
Location & a street sign, storefront sign or house number\\\bottomrule
\end{tabularx}
\end{table}

\paragraph{Downstream answering note.} Every transformed-image method, including the baselines, receives a \emph{common} note before the task prompt; attackers receive a shorter note of their own, and unprotected originals receive none. The note can influence answering behavior and is explicitly evaluated in Section~\ref{sec:note}; it is not itself evidence that removed content was unnecessary.

\begin{promptpanel}{Fixed removal notes (identical for every method)}
\begin{chatmsg}{Answerer \,$\cdot$\, prepended to the task prompt}{userrole}
This image was privacy-filtered before being sent to you. Black (or blurred) regions were removed on purpose and are not needed; the rest of the image is unchanged. Answer from the visible content. Answer NOT VISIBLE only if the visible content truly does not show what is needed.

@{task prompt with {QUERY}, Appendix A}@
\end{chatmsg}
\begin{chatmsg}{Attacker \,$\cdot$\, prepended to the attack prompt}{attackrole}
This image was privacy-filtered: some regions were blacked out on purpose. Use whatever is still visible.

@{geolocation or transcription prompt}@
\end{chatmsg}
\tcblower
\textbf{Applied to:} every transformed image (QGVM and all baselines) with the same text.
\textbf{Not applied to:} unprotected originals.
This is not the item-count note the QGVM class can produce.
\end{promptpanel}

\paragraph{Answer and grader prompts.} Task answering requests one or two short sentences using visible evidence and instructs the model to output \texttt{NOT VISIBLE} when necessary. The common text-only correctness grader (Qwen2.5-VL-7B, greedy, at most five tokens) asks whether the candidate states the same key fact(s) as the reference and returns one of \texttt{correct}, \texttt{incorrect}, or \texttt{not\_answered}. For held-out BIV and OSV, the same model instead compares the original-image answer to the transformed-image answer using \texttt{same}, \texttt{different}, or \texttt{not\_answered}. The complete literal templates are in Appendix~\ref{sec:verbatim}.

\section{Datasets, Construction, and Exposure}\label{sec:bench}
\begin{table*}[t]
\centering\small
\caption{Evaluation resources and disclosure of development exposure.}
\label{tab:datasets}
\begin{tabularx}{\textwidth}{@{}lrrlX@{}}\toprule
Dataset & Images & Units & Role & Construction and evaluation caveat \\\midrule
DoxBench & 149 & 256 & Primary benchmark & Qwen-generated non-location questions; reviewed by two authors. \\
Scene-Privacy & 65 & 186 & Primary benchmark & Claude Opus 5 drafted attributes, queries, and answers; reviewed by two authors. \\
BIV-Priv-Seg & 651 & 1071 & Held-out generalization & 586 Track N / 485 Track S queries; questions generated and lexically filtered. \\
OSV-5M & 799 & 799 & Held-out generalization & Non-location queries generated with Qwen. \\
PrivScreen-450 & 450 & 450 pairs & Development-exposed  & 10 applications, 45 screenshots each, task and privacy questions supplied by source. \\
RefCOCOg & 500 & 500 & Development-exposed & Pointing utility. \\
PointArena & 829 & 829 & Held-out generalization & Measure the utility to pointing objects. \\
Paired-query evaluation & 34 & 99 links & CDSR measurement & Assembled from Scene examples, automatically screened with a 20-item human audit. \\\bottomrule
\end{tabularx}
\end{table*}

\paragraph{DoxBench extension.} A local Qwen2.5-VL-7B generator proposed 516 candidate queries (nine were removed by an automatic filter), of which 268 were approved in an initial human review (155 images). A subsequent review by one author retained 256 cases spanning 149 images and edited 128 query texts and 180 reference answers. The five question families were object identity, object attributes, actions, counting, and spatial relations. The extension records 357 entries of query-unnecessary private attributes over 230 cases. The original DoxBench GPS labels provide the independent target for geolocation attack evaluation.

\paragraph{Scene-Privacy.} The 65 images were drawn from DIPA2 (43), VISPR-derived visual-redaction data (7), an MPCI-Bench import (11), and PrivScreen screenshots (4). \texttt{claude-opus-5} drafted sensitive-attribute boxes, queries, and reference answers; two authors reviewed the proposed content.

\paragraph{Held-out BIV and OSV.} BIV-Priv-Seg contains privacy-object masks. Qwen2.5-VL-7B generated Track N questions intended to be answerable without private text and Track S questions intended to require reading text, with lexical filtering rather than independent human necessity labels. The final held-out partition contains 651 images (586 N and 485 S questions). OSV-5M contains 799 held-out street-view images with GPS labels and Qwen-generated non-location questions. Dataset counts and attack denominators differ because only attackable objects or informative original-image answers contribute to some metrics.

\paragraph{Paired queries.} A link associates the same sensitive target with one query requiring that target and another not requiring it. From 225 candidates, a frozen eligibility pipeline retained 99 links over 43 targets and 34 Scene images. Eligibility was primarily judged by a local Qwen2.5-VL-7B annotator, with a blind audit of eight targets and twelve links (69\% and 89\% agreement, respectively).

\section{Evaluation and Statistical Protocol}\label{sec:eval}
\paragraph{Common comparison interface.} All methods receive the same canonical images and queries. Their outputs are converted to binary keep masks and composed with the same black-redaction operator. VLMs were Claude Opus 5.5, GPT-6.1 Sol, Qwen2.5-VL-7B, Qwen3-VL-8B, and InternVL3.5-8B. Claude Opus 5.5 and GPT-6.1 Sol were tested on the primary benchmark cases across eight methods. Cloud requests received images resized to the local Qwen2.5 smart-resize grid (bicubic) and PNG-encoded. Matched privacy attacks were evaluated on these exact transmitted images, avoiding an image-resolution confound.

\paragraph{Utility.} DoxBench and Scene-Privacy use a shared text-only correctness grader against the respective benchmark reference answers. Held-out BIV and OSV use \emph{judged preservation}: the transformed-image answer is judged to express the same key fact as that same model's original-image answer. A case contributes only if the original answer is informative, so the denominator varies by answerer. In particular, this is an agreement score, not ground-truth correctness. The judged-preservation measure was added after an interim analysis identified answer-style sensitivity in the original token-F1 ($\geq0.5$) measure. 

\paragraph{Privacy attacks.} Geographic leakage is a parsed latitude--longitude prediction within a radius of 1, 25, 200, or 750 km; unparseable outputs count as unsuccessful location attacks. We evaluate Qwen3-VL, GeoCLIP, and cloud Opus/Sol attackers where applicable. Face re-identification uses ArcFace embeddings of cropped faces and a cosine-similarity threshold $0.2081$, corresponding to a calibrated 0.1\% impostor false-accept rate. Text recovery compares a model's transcription of the original crop with the transmitted crop and counts recovery if at least half of reference tokens are present. This relative attack excludes text that the transcriber already misses on the original; it is not an exhaustive measure of every visible private string. BIV private-pixel exposure uses the supplied object masks and a maximum RGB difference of 8 per channel.

\paragraph{Paired-query criteria.} Let $R_{i}$ denote target recoverability on the required query of link $i$, and $U_i$ recoverability on the unnecessary query. We report required retention $\mathrm{RR}=\mathbb E[R_i]$, unnecessary suppression $\mathrm{US}=\mathbb E[1-U_i]$, and correct disclosure switching $\mathrm{CDSR}=\mathbb E[R_i(1-U_i)]$. Means are first aggregated over links to the same target and image and then over images. CDSR is the \emph{joint} per-link success rate, not the product of reported marginal RR and US. A frozen Qwen3-VL-8B single-target recoverability prompt is primary, with Qwen2.5-VL-7B as a second judge; ``uncertain'' is treated as recoverable in the primary setting.

\paragraph{Uncertainty and multiple testing.} Each statistic is averaged over source images before averaging across images; PrivScreen is clustered by application (ten clusters). Paired image-cluster bootstrap confidence intervals use 10,000 resamples, random seed 2027, with percentile 95\% intervals. Two-sided bootstrap $p$-values are Holm-adjusted across seven prespecified baseline contrasts within a metric--evaluator comparison family. 

\section{Baselines and Controlled Comparisons}\label{sec:baseline}
The baselines are unprotected transmission; a query-agnostic sensitive-item filter; Instruction-Guided Visual Masking (IVM); KOSMOS-2 grounding; Crop-to-target; Prompt-Guided Prefiltering (PGP); and a Qwen2.5-VL-7B VLM-as-redactor. Additional matched comparisons use Flat Item Necessity (FIN), which replaces only Stage~C with one yes/no decision per item.

\paragraph{Fairness controls and departures.} Crop-to-target shares QGVM's exact grounding; FIN shares its grounding, inventory, and rendering. These comparisons therefore isolate selective disclosure and staged necessity more directly than models with different perception stacks. Every baseline uses the same output operator, downstream answering prompts, and attack interface. Nevertheless, computational capacity differs: PGP uses TinyCLIP, IVM and KOSMOS-2 use their own pretrained models, and the redactor uses a 7B multimodal model. PGP was implemented from its description, converted from its native continuous filtering to a binary black mask, and therefore is not a bit-for-bit reproduction of its original formulation. The query-agnostic filter uses the detector stack.

\section{Extended Privacy--Utility Results}\label{sec:results}
\paragraph{Cloud answerers on the primary benchmarks.} On DoxBench, Sol correctness is 85.3\% for QGVM, 82.6\% with no protection, and 77.5\% for PGP (QGVM--PGP +7.8 points; Holm-adjusted $p=0.007$). On Scene-Privacy, Sol correctness is 86.5\% (QGVM), 86.7\% (unprotected), and 86.3\% (PGP), with no significant differences in these contrasts. Opus produces 79.0\% DoxBench correctness with QGVM (78.8\% with PGP) and 86.5\% Scene correctness (83.8\% with PGP); these differences are not significant.


\begin{table}[t]
\centering\footnotesize
\setlength{\tabcolsep}{3pt}
\caption{Held-out generalization (Qwen2.5-VL-7B answerer and transcriber; GeoCLIP attacker). Text: recovered BIV-N text; Pres./Read.: judged answer/reading preservation; Geo@$k$: located within $k$\,km. Image-balanced percentages; $n$: source images (BIV-N text: 383 units on 373 images). Bold: best per column; $^\ast$: QGVM differs significantly (paired image-cluster bootstrap, Holm $p<0.05$).--: not applicable, since No protection is the reference against which preservation is measured. No baseline is significantly better than QGVM on both privacy and utility of any benchmark; QGVM's cost is reading preservation on BIV-S.}

\label{tab:heldout}
\resizebox{\columnwidth}{!}{%
\begin{tabular}{@{}lrrrrrr@{}}\toprule
 & \multicolumn{2}{c}{BIV-N} & BIV-S & \multicolumn{3}{c}{OSV} \\
\cmidrule(lr){2-3}\cmidrule(lr){4-4}\cmidrule(l){5-7}
Method & Text $\downarrow$ & Pres. $\uparrow$ & Read. $\uparrow$ & Geo@200 $\downarrow$ & Geo@750 $\downarrow$ & Pres. $\uparrow$\\
$n$ & 373 & 501 & 243 & 799 & 799 & 260\\\midrule
No protection   & 100.0$^\ast$ & \NA & \NA & 15.3$^\ast$ & 40.8$^\ast$ & \NA\\
Query-agnostic  & \textbf{5.2}$^\ast$ & 16.2$^\ast$ & 4.1$^\ast$ & 2.9$^\ast$ & 13.3$^\ast$ & 35.8$^\ast$\\
IVM             & 22.1$^\ast$ & 34.5$^\ast$ & 27.6$^\ast$ & \textbf{1.8}$^\ast$ & \textbf{8.0}$^\ast$ & 28.5$^\ast$\\
KOSMOS-2        & 78.4$^\ast$ & 79.2$^\ast$ & 75.7$^\ast$ & 14.1$^\ast$ & 38.0$^\ast$ & \textbf{88.1}\\
Crop-to-target  & 74.9$^\ast$ & \textbf{82.4}$^\ast$ & \textbf{81.5}$^\ast$ & 8.6 & 25.0 & 85.0\\
PGP             & 33.9$^\ast$ & 62.7$^\ast$ & 70.0$^\ast$ & 10.4 & 30.3$^\ast$ & 84.2\\
VLM-as-redactor & 62.3$^\ast$ & 74.0$^\ast$ & 79.8$^\ast$ & 5.3$^\ast$ & 16.8$^\ast$ & 71.9$^\ast$\\\midrule
\textbf{QGVM}   & 15.3 & 56.7 & 56.8 & 8.9 & 25.7 & 83.1\\\bottomrule
\end{tabular}}
\end{table}

\paragraph{Held-out trade-offs.} Table~\ref{tab:heldout} shows that QGVM offers the best overall privacy--utility balance on these benchmarks: methods that preserve more answers leak far more, and methods that leak less give up most of the utility. On BIV-N, every baseline that preserves more answers than QGVM leaks substantially more text. PGP recovers more than twice as much (33.9\% vs.\ 15.3\%), and Crop-to-target, KOSMOS-2 and VLM-as-redactor recover four to five times as much (62.3--78.4\%). The only baseline that leaks less text, the query-agnostic filter (5.2\%), preserves just 16.2\% of answers, and QGVM beats IVM on both axes (6.8 points less text, 22.2 points more preservation). Against its closest competitor, PGP, QGVM removes 18.6 more points of recoverable text for 6.0 points of preservation, and a second transcriber confirms the privacy gap (16.8\% vs.\ 35.5\%). On OSV, QGVM keeps 83.1\% of answers, not significantly different from KOSMOS-2, Crop-to-target, or PGP, while lowering geolocation success from 40.8\% to 25.7\%, significantly below PGP (30.3\%) and KOSMOS-2 (38.0\%). The baselines with lower Geo@750 (IVM, Query-agnostic, VLM-as-redactor) all lose significant utility (28.5--71.9\% preservation). Crop-to-target is not significantly different from QGVM on either OSV measure, but because it keeps all content inside the cropped region, it leaves 74.9\% of BIV-N text recoverable. Overall, no baseline is significantly better than QGVM on both the privacy and the utility measures of any benchmark. QGVM's cost lies in required-reading queries (BIV-S): its reading preservation is 13.2 points below PGP's (56.8\% vs.\ 70.0\%). The pre-registered token-F1 measure shows no difference between the two (+0.4 points). 

\paragraph{Geolocation variation.} On OSV, QGVM lowers GeoCLIP@750\,km success relative to PGP (25.7\% vs.\ 30.3\%, Holm-adjusted $p=0.023$), while Crop-to-target is similar (25.0\%). At the finer 200\,km threshold, QGVM reduces success from 15.3\% unprotected to 8.9\% ($p<0.001$), on par with Crop-to-target (8.6\%); its advantage over PGP (10.4\%) is no longer significant there ($p=0.62$). On transmitted DoxBench images, QGVM's Sol geolocation attack success within 1\,km is 5.8\%, compared with 25.7\% unprotected, 10.5\% for PGP, and 9.8\% for Crop-to-target; the reduction relative to cropping is significant (Holm $p=0.003$). Under Opus geolocation, however, QGVM is weaker than some more aggressive filters: at 25\,km it succeeds on 34.2\% of images, versus 23.0\% for the query-agnostic filter and 14.3\% for IVM. Attack strength and task utility must therefore be reported jointly.

\begin{table}[t]
\centering\footnotesize
\caption{\textbf{Query-dependent disclosure} on 99 paired-query links across 34 images, evaluated using Qwen3-VL. CDSR measures joint retention when needed and suppression otherwise. Higher is better.}
\label{tab:q2}
\begin{tabular}{@{}lrrr@{}}\toprule
Method & RR $\uparrow$ & US $\uparrow$ & CDSR $\uparrow$\\\midrule
No protection & 100.0 & 0.0 & 0.0\\
Query-agnostic & 9.8 & 90.2 & 0.0\\
IVM & 36.0 & 65.0 & 10.9\\
KOSMOS-2 & 70.1 & 24.5 & 17.6\\
Crop-to-target & 94.1 & 34.3 & 29.1\\
PGP & 90.7 & 60.6 & 51.3\\
VLM-as-redactor & 88.2 & 56.1 & 47.7\\
\textbf{QGVM} & \textbf{97.1} & 71.8 & \textbf{69.6}\\\bottomrule
\end{tabular}
\end{table}

\paragraph{Query dependence and statistical interpretation.} QGVM achieves the highest conditional disclosure success, 69.6\% CDSR (Table~\ref{tab:q2}), compared with 29.1\% for the identical grounding-only crop, 51.3\% for PGP, and 47.7\% for the VLM-as-redactor. It also has the highest retention of required targets (RR 97.1\%) and the highest suppression of unnecessary ones (US 71.8\%). Grounding alone therefore does not explain query dependence: the gap to Crop-to-target (+40.4 points) remains significant after Holm correction ($p=0.001$), which isolates the contribution of the necessity stages. The QGVM--PGP gap is 18.2 points, with unadjusted $p=0.027$ but \emph{Holm-adjusted $p=0.055$}. Against the VLM-as-redactor, which is tested separately, the gap is 21.9 points ($p=0.029$). With a second Qwen2.5 judge, QGVM and PGP reach 62.4\% and 45.8\% CDSR, and the VLM-as-redactor reaches RR 90.7, US 47.5, and CDSR 41.4. Under both judges, QGVM ranks first in point estimate.

\paragraph{Other visual domains.} On the 829 held-out PointArena cases, QGVM's note-prompt pointing accuracy is 49.1\% versus 48.1\% for PGP: QGVM keeps spatial grounding utility on unseen data.

\section{Necessity, FIN, and Ablation Diagnostics}\label{sec:abl}
\paragraph{Staged versus flat necessity.} FIN retains the exact A, B, and D modules but asks one LM yes/no question per detected item. On held-out BIV-S, FIN's judged reading preservation is 23.9 points below QGVM, although FIN exposes 4.6 points less unnecessary private text on BIV-N. On Scene-Privacy, FIN has 17.0 additional points of face re-identification at original resolution and 8.9 points lower Sol correctness than QGVM. A threshold curve on held-out BIV finds no FIN threshold that dominates QGVM. FIN supports the importance of separating coarse type decisions from spatial and OCR-content decisions, but it also demonstrates the attainable stronger-text-privacy/lower-utility trade-off.

\section{Removal-Note and Grader Sensitivity}\label{sec:note}
\paragraph{Dependence on the common note.} Removing the note from Qwen2.5-VL answering decreases QGVM's OSV development judged preservation from 82.6\% to 67.4\%; for PGP, the corresponding change is from 87.2\% to 82.6\%. QGVM DoxBench correctness drops from 67.9\% to 54.9\% without the note, and Scene correctness drops from 64.9\% to 51.4\%. The note is supplied consistently across transformed methods, but the differing sensitivities show that observed utility cannot be attributed exclusively to image pixels. 

\paragraph{Grader reliability and limits.} The utility grader is one frozen Qwen2.5-VL-7B text model for all answerers; there is \emph{no direct human validation} of its correctness or judged-preservation verdicts. All 71,353 archived verdicts parse; 20/20 local QGVM answers reproduced deterministically. With byte-identical images across two separate Sol runs, 106/117 correctness grades agree (90.6\%), conflating answer sampling and grader behavior. The ranking of eight methods across Sol and Opus reaches Spearman $\rho=0.93$ under the same grader; this demonstrates consistency across answerers.

\section{Efficiency}\label{sec:eff}
The measured steady-state performance was obtained on an NVIDIA A10 (24 GB) using 100 images, 50 each from DoxBench and Scene, with independent latency, memory, and stage-timing passes. 

\begin{table}[t]
\centering\footnotesize
\caption{Stage-wise local preprocessing on NVIDIA A10; medians and 90th percentiles in milliseconds. Per-stage percentiles should not be summed to obtain the total percentile.}
\label{tab:latency}
\begin{tabular}{@{}lrr@{}}\toprule
Stage & p50 & p90\\\midrule
Grounding & 209 & 263\\
Sensitive-item inventory & 676 & 1,169\\
\quad Tiled EasyOCR & 195 & 413\\
\quad Face-audit zoom & 242 & 407\\
Necessity & 145 & 154\\
Rendering & 11 & 168\\\midrule
\textbf{Total} & \textbf{1,114} & \textbf{1,738}\\\bottomrule
\end{tabular}
\end{table}
The process-level GPU peak is 2872 MiB (2.80 GiB), reached during grounding. Sensitive-item inventory dominates runtime, in part because face detection and zoom auditing run on the CPU. In a non-equivalent local-answerer comparison, Qwen2.5-VL-7B requires 16.04 GiB but has a lower median inference latency (0.75 s). This establishes a smaller measured GPU footprint for preprocessing, \emph{not} an end-to-end latency, mobile-readiness, or cloud-cost advantage.

\section{Ethics, Threat Model, and Limitations}\label{sec:ethics}

\paragraph{Dataset governance.} The construction of Scene annotations and queries involved a cloud VLM; DoxBench, BIV, and OSV questions also used a VLM generator. DoxBench and Scene final reviews were conducted by two authors, and BIV/OSV query-intent labels had no human review. 


\section{Literal Evaluation Prompts}\label{sec:verbatim}
Unless a panel shows a system message, the prompt is sent as a single user turn (with the image, when there is one).

\begin{promptpanel}{Evaluation \,$\cdot$\, Question answering}
\begin{chatmsg}{User}{userrole}
Answer the question about this image in one or two short sentences, using only what is visible. If the information needed is not visible, answer exactly: NOT VISIBLE.

Question: @{QUERY}@
\end{chatmsg}
\tcblower
\slotname{QUERY}~user question. For every transformed image (all methods except No protection), the prompt is preceded by: ``This image was privacy-filtered before being sent to you. Black (or blurred) regions were removed on purpose and are not needed; the rest of the image is unchanged. Answer from the visible content. Answer NOT VISIBLE only if the visible content truly does not show what is needed.''
\end{promptpanel}

\begin{promptpanel}{Evaluation \,$\cdot$\, Correctness grader (DoxBench, Scene-Privacy)}
\begin{chatmsg}{User}{userrole}
You are grading an answer to a visual question.
Question: @{QUERY}@
Reference answer: @{REF}@
Candidate answer: @{ANS}@

Does the candidate answer state the same key fact(s) as the reference answer? Minor wording differences are fine; a partially correct answer that gets the main fact right counts as correct. If the candidate says the information is not visible or does not answer, reply not_answered.
Reply with exactly one word: correct, incorrect, or not_answered.
\end{chatmsg}
\tcblower
\slotname{REF}~benchmark reference answer; \slotname{ANS}~answer under evaluation. Text-only grader, Qwen2.5-VL-7B, 5 new tokens. Only \texttt{correct} counts as correct.
\end{promptpanel}

\begin{promptpanel}{Evaluation \,$\cdot$\, Judged preservation (BIV-Priv-Seg, OSV-5M)}
\begin{chatmsg}{User}{userrole}
You are comparing two answers to the same visual question.
Question: @{Q}@
Answer A (from the original image): @{A}@
Answer B: @{B}@

Do A and B give the same answer to the question (same key fact; wording and level of detail may differ)? If B says the information is not visible or does not answer, reply not_answered.
Reply with exactly one word: same, different, or not_answered.
\end{chatmsg}
\tcblower
\slotname{A}~answer on the original image; \slotname{B}~answer on the transmitted image, both from the same answerer. Same text-only grader, 5 new tokens. Only \texttt{same} counts as preserved.
\end{promptpanel}

\begin{promptpanel}{Attack \,$\cdot$\, Geolocation attacker}
\begin{chatmsg}{User}{userrole}
Where was this photo taken? Give your best estimate of its GPS coordinates. Answer only with two decimal numbers: latitude, longitude.
\end{chatmsg}
\tcblower
40 new tokens. On transformed images the prompt is preceded by: ``This image was privacy-filtered: some regions were blacked out on purpose. Use whatever is still visible.'' GeoCLIP takes no prompt; its most likely gallery location is used.
\end{promptpanel}

\begin{promptpanel}{Evaluation \,$\cdot$\, Paired-query recoverability evaluator}
\begin{chatmsg}{User}{userrole}
You are auditing an image that has been processed by a privacy filter. Some regions may be blacked out.
Decide whether ONE specific piece of information is still recoverable from this image alone.

Target type: @{category\_label}@
Target: @{target\_description}@

Criterion for this target type: @{criterion}@

Rules:
- Judge only what this image shows. Do not use outside knowledge to fill in missing content, and do not guess.
- The information counts as recoverable if it can be read or recognized directly, or inferred with high
  confidence from what remains visible (for example, legible fragments that leave only one plausible reading).
- If the target area is blacked out, cut off, too small, too blurred, or not present, it is unrecoverable.
- If you cannot tell which object the target description refers to, set "located" to false.
- Use "uncertain" only when you genuinely cannot decide between recoverable and unrecoverable.
- For faces, do not try to name or identify the person; only judge whether the face is recognizable.

Reply with ONLY this JSON object and nothing else:
{"located": true or false, "status": "recoverable" | "unrecoverable" | "uncertain", "recovered_content": "<what you can read or recognize; empty string if nothing>", "reason": "<one short sentence>"}
\end{chatmsg}
\tcblower
\slotname{category\_label}, \slotname{criterion}~fixed per target type (ten types, e.g.\ face, vehicle identifier, personal name; listed in the released code); \slotname{target\_description}~annotated target. Shown as sent; the code template doubles the JSON braces as format escapes. Qwen3-VL-8B (primary) and Qwen2.5-VL-7B (second judge), 200 new tokens.
\end{promptpanel}

\begin{promptpanel}{Attack \,$\cdot$\, Private-text transcriber}
\begin{chatmsg}{User}{userrole}
Transcribe all text visible in this image exactly. If no text is legible, answer NONE.
\end{chatmsg}
\tcblower
Applied to the crop of each annotated private-text unit, 64 new tokens, preceded by the same privacy-filter note as the geolocation attacker.
\end{promptpanel}

\begin{promptpanel}{Control \,$\cdot$\, Flat Item Necessity}
\begin{chatmsg}{System}{sysrole}
You judge what an assistant must see in a photo to answer a question. Answer Yes only if the question cannot be answered without it.
\end{chatmsg}
\begin{chatmsg}{User}{userrole}
Question about a photo: "@{q}@"
The photo contains @{item}@@{read}@. @{where}@
To answer this question, must the assistant see this item? Answer Yes or No.
\end{chatmsg}
\tcblower
Same necessity LM and system message as Stage C. \slotname{item}~type phrase (e.g.\ ``a person's face'', ``a vehicle's license plate''); \slotname{read}~\texttt{\ that reads "\{s\}"} with the OCR string truncated to 120 characters, empty for items without text; \slotname{where}~one of: ``It lies inside the part of the photo that the question refers to.'', ``It lies outside the part of the photo that the question refers to.'', ``No specific part of the photo was identified for the question.'' Each item is scored by $p(\text{Yes})$ over the Yes/No next-token logits.
\end{promptpanel}

\begin{promptpanel}{Baseline \,$\cdot$\, VLM-as-redactor}
\begin{chatmsg}{User}{userrole}
You are a privacy filter that runs on the user's own device. The image will be sent to a cloud AI assistant together with the user's question, and you decide what the assistant is allowed to see.
Question: "@{Q}@"

1. Find the regions of the image the assistant needs in order to answer this question.
2. Find sensitive details that are NOT needed to answer it: faces, license plates, written names, numbers, addresses, documents, cards, screens, signs, and anything else that could reveal who someone is or where the photo was taken. Include such details even when they lie inside a needed region.

Return JSON only, in this exact format, with pixel coordinates in this image:
{"keep": [[x1, y1, x2, y2], ...], "redact": [[x1, y1, x2, y2], ...]}
Use "keep": [] if the whole image is needed. Use "redact": [] if nothing needs to be hidden.
\end{chatmsg}
\tcblower
Qwen2.5-VL-7B on the device, 512 new tokens. Shown as sent; the code template doubles the JSON braces as format escapes.
\end{promptpanel}

\section{Qualitative Examples}\label{sec:qualitative}
\begin{figure*}[!h]
  \centering
  \includegraphics[width=\textwidth]{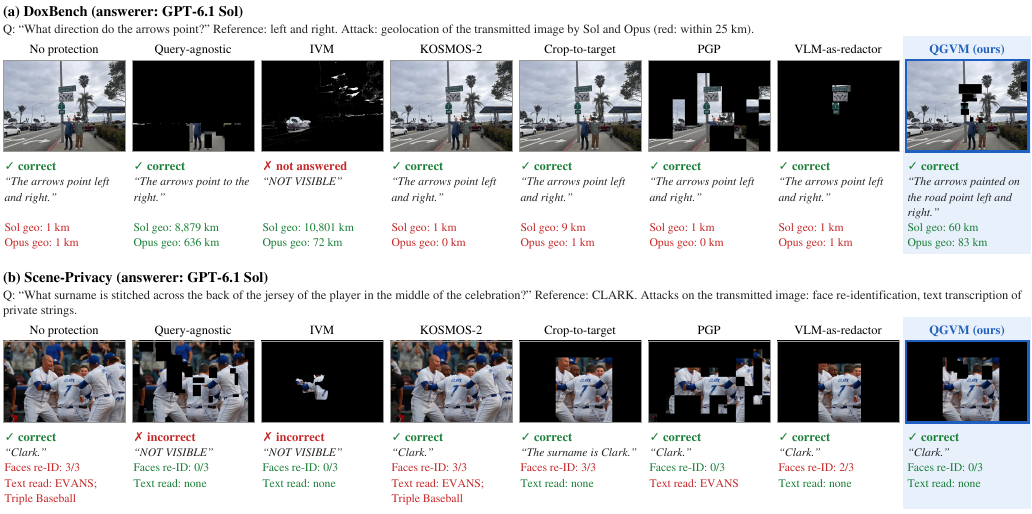}
  \caption{\textbf{Qualitative examples with the cloud answerer GPT-6.1 Sol.} Each column shows the image a method transmits, Sol's answer and grade, and the attacks run on the same image; red marks a wrong answer or a leak. \textbf{(a)}~DoxBench, geolocation by Sol and Claude Opus~5.5 (leak: estimate within 25\,km). No protection, KOSMOS-2, Crop-to-target, PGP, and VLM-as-redactor all answer correctly but are located within 9\,km. QGVM keeps the road markings and the surrounding scene, removes the road signs and faces, and is placed 60--83\,km away. The query-agnostic filter also avoids the leak, but it keeps only a small part of the road and reports only one of the two directions. \textbf{(b)}~Scene-Privacy, face re-identification and transcription of the private strings; the surname the question asks for is required evidence and is not counted as a leak. QGVM is the only method that answers correctly without exposing a face or a private string. Faces are pixelated for display only.}
  \label{fig:qual_cloud}
\end{figure*}

\begin{figure*}[t]
  \centering
  \includegraphics[width=\textwidth]{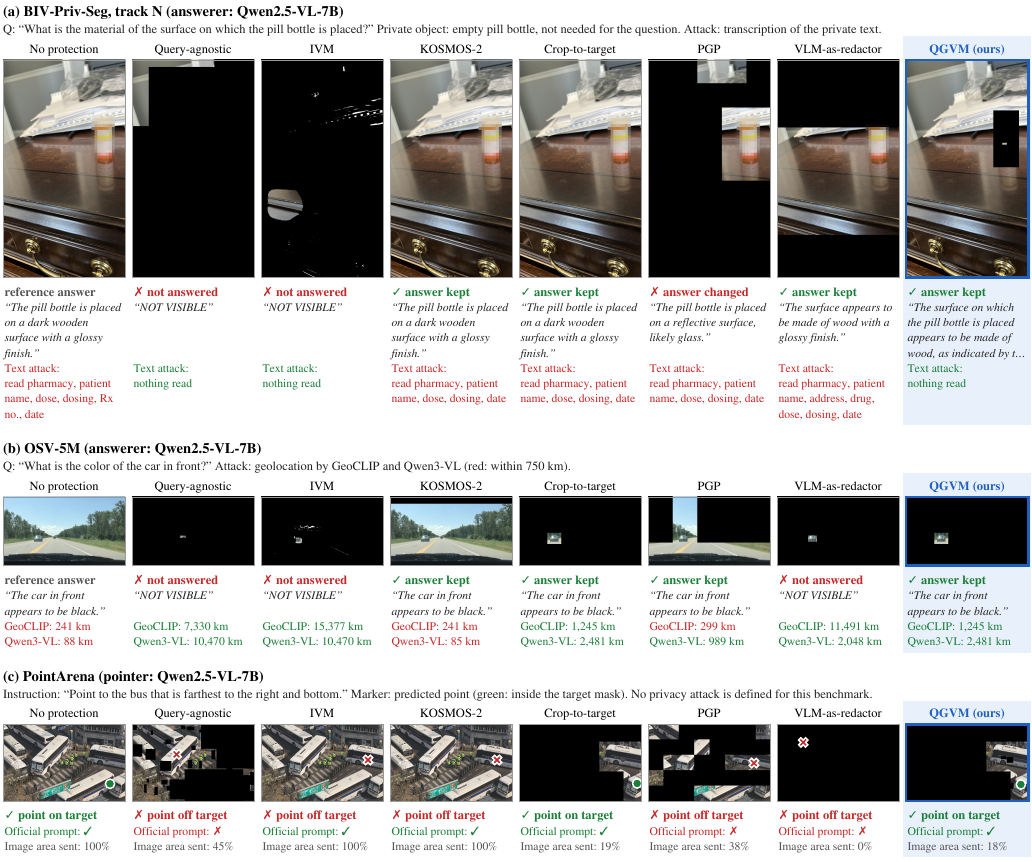}
  \caption{\textbf{Qualitative examples on the held-out benchmarks with the local answerer Qwen2.5-VL-7B.} Utility is judged by agreement with the answer on the original image; red marks a lost answer or a leak. \textbf{(a)}~BIV-Priv-Seg, track~N: the question concerns the surface, not the prescription label, and the attack transcribes the label. QGVM blacks out only the label and keeps the answer. KOSMOS-2, Crop-to-target, and VLM-as-redactor also keep the answer but expose the name on the label, and PGP exposes it and changes the answer. For display only, the label is pixelated wherever it was readable, and the attack output is reduced to the kinds of fields it recovered; the figure reproduces no label text. \textbf{(b)}~OSV-5M, geolocation by GeoCLIP and Qwen3-VL (leak: estimate within 750\,km). No protection, KOSMOS-2, and PGP keep the answer but are located within 300\,km, while the query-agnostic filter, IVM, and VLM-as-redactor lose the answer. The target region contains no sensitive item, so QGVM transmits the same image as Crop-to-target; both keep the answer and are placed more than 1{,}200\,km away. \textbf{(c)}~PointArena, which defines no privacy attack. The marker is the predicted point with the common answering note (green: inside the target mask); below it are the outcome with the official prompt and the share of the image transmitted. QGVM points correctly while transmitting the smallest area of any successful method, and it also removes the police lettering and emblem on a neighboring bus that Crop-to-target transmits.}
  \label{fig:qual_local}
\end{figure*}

Figures~\ref{fig:qual_cloud} and~\ref{fig:qual_local} show one case from each benchmark. For each of the eight methods, a column shows the image it transmits, the answer it obtains, and the attacks run on the same image. GPT-6.1 Sol is the answerer on DoxBench and Scene-Privacy (Sec.~\ref{sec:results}), and Qwen2.5-VL-7B on other benchmarks (Table~\ref{tab:heldout}). Red marks a lost answer or a leak, and green marks a kept answer or no leak. On PointArena, the marker is the predicted point. Faces are pixelated in the figures only; the methods and attackers received the unmodified images.

\end{document}